\documentclass[10pt,twocolumn,letterpaper]{article}

\usepackage[pagenumbers]{cvpr} 

\usepackage{url}
\usepackage{lineno}
\usepackage{graphicx}
\usepackage{amssymb,amsmath}
\usepackage{xcolor}
\usepackage{bbding}
\usepackage{multirow}
\usepackage{subcaption}

\newcommand{\blue}[1]{\textcolor{blue}{#1}}
\newcommand{\red}[1]{\textcolor{red}{#1}}

\newcommand\Tstrut{\rule{0pt}{2.6ex}}         
\newcommand\Bstrut{\rule[-1.0ex]{0pt}{0pt}}   

\newcommand{\Sref}[1]{Section \ref{#1}}
\newcommand{\Tref}[1]{Table \ref{#1}}
\newcommand{\Fref}[1]{Figure \ref{#1}}

\usepackage[pagebackref,breaklinks,colorlinks]{hyperref}

\def\cvprPaperID{*****} 
\def\confName{CVPR}
\def\confYear{2022}

\title{Efficient Deep Neural Network for Photo-realistic Image Super-resolution}

\author{
Namhyuk Ahn\textsuperscript{1,3} \quad Byungkon Kang\textsuperscript{2} \quad Kyung-Ah Sohn\textsuperscript{3} \\
{\textsuperscript{1} NAVER WEBTOON AI \quad
\textsuperscript{2} SUNY Korea \quad
\textsuperscript{3} Ajou University}
}

\begin{document}
\maketitle

\begin{abstract}
Recent progress in deep learning-based models has improved photo-realistic (or perceptual) single-image super-resolution significantly.
However, despite their powerful performance, many methods are difficult to apply to real-world applications because of the heavy computational requirements.
To facilitate the use of a deep model under such demands, we focus on keeping the network efficient while maintaining its performance.
In detail, we design an architecture that implements a cascading mechanism on a residual network to boost the performance with limited resources via multi-level feature fusion.
In addition, our proposed model adopts group convolution and recursive schemes in order to achieve extreme efficiency. We further improve the perceptual quality of the output by employing the adversarial learning paradigm and a multi-scale discriminator approach. The performance of our method is investigated through extensive internal experiments and benchmarks using various datasets. Our results show that our models outperform the recent methods with similar complexity, for both traditional pixel-based and perception-based tasks.
\end{abstract}

\section{Introduction}
\label{sec:intro}
Image super-resolution (SR) is a longstanding computer vision task that can be widely used in many applications.
This task focuses on recovering a high-resolution (HR) image from low-resolution (LR) images.
In particular, single-image super-resolution (SISR) performs SR using a single LR image.
Since the SISR problem is a one-to-many mapping, constructing an effective SISR algorithm is challenging.
Despite the difficulties, SISR has been actively studied since it can be applied to a variety of scenarios (\textit{i.e.,} face~\cite{huang2010super,chan2021glean} or biometrics~\cite{nguyen2018super}).
Recently, deep learning-based methods have shown prominent performance on SR task~\cite{srcnn2014,liang2021swinir,zhang2021designing,wang2021real}.
The major trend of deep models is not only stacking layers to their networks~\cite{mdsr2017} but also designing and assembling internal blocks and network topologies~\cite{srdense} to achieve more accurate results.

Even though SR performance continues to improve, there still exists a gap between the quantitative scores and human-perceived judgment.
Various methods adopt pixel-based (or distortion-based) error functions, such as mean squared error (MSE) or L1 loss, to train the SR network.
Minimizing such objectives leads to a high peak signal-to-noise ratio (PSNR) score, which is a commonly used quality measure in the SR community.
However, the ability to restore the high-frequency details in such cases is limited, since pixel-based error functions only capture the difference between two images pixel-wise.
Moreover, they often result in blurry output images, thus usually disagreeing with the subjective evaluation scores given by human judges.
To address such shortcomings, several deep learning-based methods perceptually optimize their network to improve human-visual quality.
Starting with SRGAN~\cite{srgan}, most of the models that aim for good perceptual quality employ the generative adversarial network (GAN)~\cite{goodfellow2014generative} paradigm and perceptual loss~\cite{johnson2016perceptual}.
Enhanced SRGAN (ESRGAN)~\cite{wang2018esrgan} achieves the best perceptual quality by improving both the generator and the discriminator simultaneously.

Although deep learning-based networks significantly increase the quality of the SR outputs, applying such models to real-world scenarios is another challenge.
There are many cases that require not only quality but also efficiency such as streaming services or mobile applications.
However, the recently proposed methods use deep networks, which can be computationally heavy.
From this perspective, designing a lightweight SR network is very crucial.

\begin{figure*}[t]
\centering
\includegraphics[width=\linewidth]{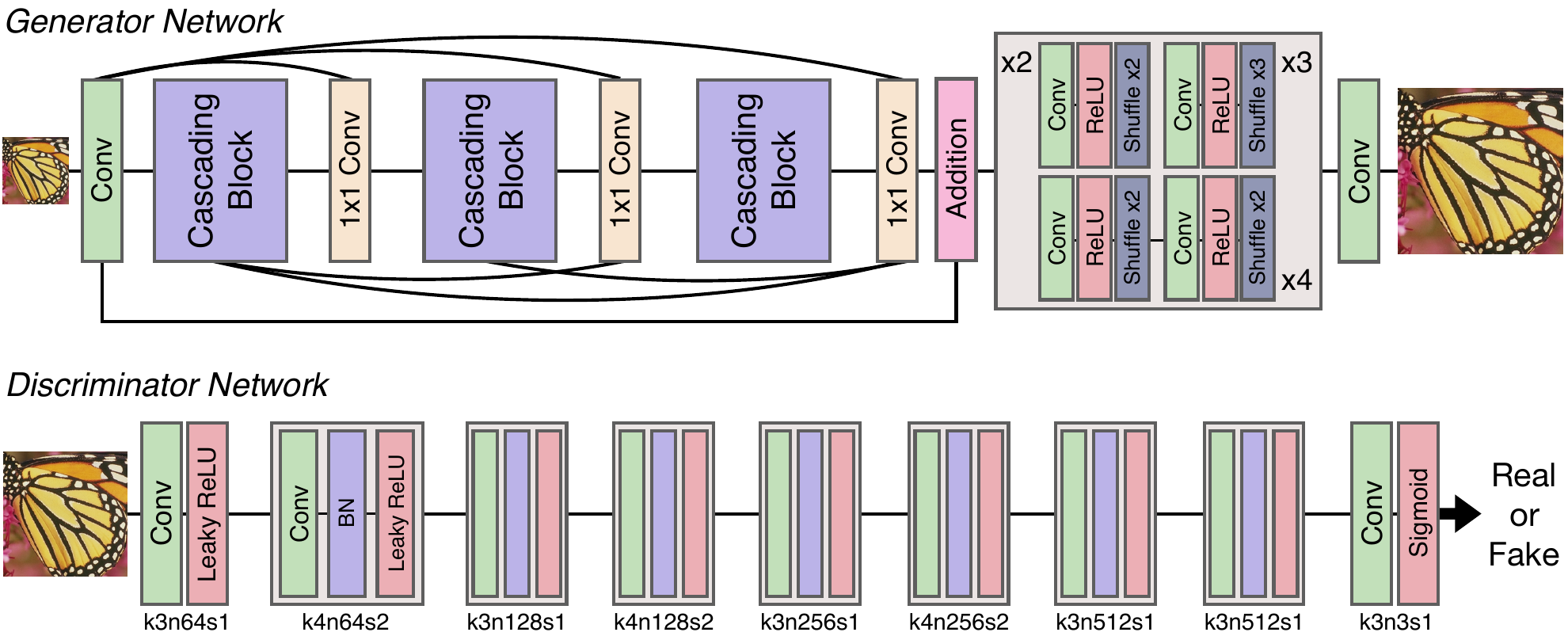}
\caption{\textbf{Network architecture.} (top) Generator network. This network consists of cascading blocks and upsample blocks. (bottom) Discriminator network with corresponding kernel size ($k$), number of feature map ($n$) and stride ($s$) indicated for each convolution layer.}
\label{fig:model}
\end{figure*}

Several works~\cite{kim2016deeply,memnet,MSLapSRN} make efforts to design a \textit{lightweight} SR model by reducing the number of parameters.
One of the most simple and effective approaches is to construct the model in a recursive manner~\cite{kim2016deeply}.
However, even though such studies show good SR performance using a small number of parameters, they have some downsides:
These works increase the depth or width of the network to compensate for the performance loss caused by the use of the recursive scheme, making the inference very slow. 
Moreover, their \textit{early-upsample} design, which upsamples the input image before inputting it to the network, results in high computational cost.

However, as mentioned earlier, the number of operations is also an important factor to consider in real-world demands.
For the SR systems that operate on mobile devices, the execution speed also plays an important role from a user-experience perspective.
Especially the battery capacity, which is heavily dependent on the amount of computation performed, becomes a major problem.
In this respect, reducing the number of operations is a challenging and necessary step that has largely been ignored until now.
A relevant practical scenario can be found in video streaming services.
The demand for streaming media has skyrocketed, and hence large storage for massive multimedia data is required.
It is therefore imperative to compress data using lossy compression techniques before storing.
Then, an SR technique can be applied to restore the data to the original resolution.
However, because latency is the most critical factor in such services, the decompression process has to be performed in near-real time.
To do so, it is essential to make the SR methods lightweight in terms of the number of operations in order to satisfy timing constraints.

To handle these requirements and improve the recent models, we propose a photo-realistic cascading residual network (PCARN), which is an extended version of our preliminary work, cascading residual network (CARN)~\cite{ahn2018fast}.
Following the ESPCN~\cite{espcn2016}, CARN and PCARN take the LR images and compute the HR counterparts as the output of the network.
Based on this architecture, we introduce a cascading mechanism at both the \textit{local} and \textit{global} levels to incorporate features from multiple layers\footnote{Here, we use terms \textit{local} and \textit{global} to distinguish where the features are generated from. In other words, ``local'' means inside of the internal blocks, and ``global'' means the outer part of the network.}.
It has the effect of gathering various levels of intermediate feature representations in order to receive more information that can be helpful when restoring degraded images.
By extending our previous version~\cite{ahn2018fast}, we have conducted additional experiments to dissect the effectiveness of the proposed model, including the initialization strategy.
Since the CARN has several narrow 1$\times$1 convolution layers (by its design), we observed that these layers severely affect the quality of the initialization, and high-variance initial values results in degraded performance.

In addition, we extend our prior work by tackling the limitation of pixel-based loss.
The SR methods with this loss produce blurry and sometimes unsatisfactory outputs.
To overcome such an issue, we adopt adversarial training to build photo-realistic CARN (PCARN).
More specifically, we set the CARN model as a generator and attach an additional discriminator network that distinguishes whether the input images are from the HR or the SR set (\Fref{fig:model}).
Additionally, we also enhance the discriminator by using a multi-scale discriminator strategy instead of using a single discriminator to make the model produce images with high perceptual quality. 
The multi-scale discriminator consists of multiple networks, where each network is in charge of handling a certain scale.
It improves the ability of the generator and the discriminator to preserve the details by taking into account both the coarse and fine textures.
Furthermore, as our prior work~\cite{ahn2018fast}, we build the PCARN-M (mobile) to allow users to tune the trade-off between the quality and the heaviness of the model. It is implemented with the efficient residual block and a recursive network scheme.

In summary, our contributions are as follows:
\textbf{1)} We propose PCARN, a neural network model based on novel cascading modules that effectively boost the SR performance via multi-level representation and multiple shortcut connections. 
\textbf{2)} With GAN-based learning and a multi-scale discriminator, our model can capture fine details effectively.
We demonstrate that our method achieves superior performance in various image quality assessments and shows a good balance of the perception-distortion trade-off with limited resources compared to the previous methods.
\textbf{3)} We propose PCARN-M for efficient SR by combining the efficient residual block and the recursive network scheme.
Experimental results demonstrate that our method is substantially faster and more lightweight than the recent deep learning-based methods on perceptual-based SR task.

\section{Related Work}
\label{sec:related_work}
In this section, we first focus on the deep learning-based SR models (\Sref{sec:sisr}).
In \Sref{sec:real-sisr}, we discuss the recent advances on the photo-realistic SR task.
Finally, we briefly review the network compression in \Sref{sec:efficient_nn}.
\subsection{Deep Learning-based SR}
\label{sec:sisr}
The performance of the SR has been greatly improved with the powerful capabilities of the deep learning-based methods.
As a pioneer work, Dong et al.,~\cite{srcnn2014} propose a deep learning-based model, SRCNN, that surpasses the traditional approaches.
However, SRCNN requires large computation resources compared to its depth, since the model takes upsampled images as an input.
On the other hand, ESPCN~\cite{espcn2016} takes an LR image as an input, after which it upsamples the image at the end of the network. This strategy reduces the computation substantially compared to the \textit{early-upsample} scheme. 

A shortcoming of the aforementioned methods is that they only use a few convolutional layers because of training instability.
To tackle this issue, VDSR~\cite{vdsr2016} introduces global residual learning and shows significant improvement over the previous methods by stacking more layers.
The global residual learning maps the LR image $\mathbf{x}$ to its residual image $\mathbf{r}$.
Then, it produces the SR image $\mathbf{\tilde{y}}$ by adding the residual back to the original, \textit{i.e.,}, $\mathbf{\tilde{y} = x + r}$.
The ESCN~\cite{wang2017ensemble} uses an ensemble technique to overcome the training instability and to increase the representation power.
All the methods mentioned directly super-resolve to the desired spatial resolution, resulting in unsatisfying quality when the input is severely downsampled.
To tackle this issue, recent studies use a progressive upsampling~\cite{wang2018fully,ahn2018image}, which upsamples the intermediary features periodically to restore the image gradually.

One possible disadvantage of applying a deep SR method is the efficiency of the network.
That is, there is a problematic increase in the size of the model.
To address this concern, most previous studies~\cite{kim2016deeply,memnet,MSLapSRN} aim to build a lightweight model in terms of the number of parameters.
DRCN~\cite{kim2016deeply} and MemNet~\cite{memnet} use a recursive layer to boost the SR quality without additional parameters.
Similarly, MSLapSRN~\cite{MSLapSRN} ties the parameters of each scale-wise block and takes advantage of both the recursive scheme and progressive approach, resulting in superior SR performance in terms of both SR quality and efficiency.
However, many of the parameter-efficient methods use deep networks to compensate for degraded performance caused by the use of the recursive scheme, thus require heavy computing resources.
In contrast, we aim to build a model that is lightweight in both size and computational aspects.

\subsection{Photo-realistic SR}
\label{sec:real-sisr}
Generally, deep learning-based SR networks are trained using pixel-based (or distortion-based) loss functions (\textit{e.g.,} MSE or L1 loss).
The network with these objectives can be optimized easily, but it tends to create blurry artifacts and fails to recover the structural details.
This characteristic can be problematic since a human can judge the absence of high-frequency information effortlessly~\cite{srgan}.
Hence, to overcome the inherent issue of using pixel-based losses, a generative adversarial network (GAN)~\cite{goodfellow2014generative} has been adopted to the SR field~\cite{srgan}.
By doing so, GAN-based methods show promising results in preserving human-perceptive quality.
However, since using only an adversarial loss makes the training process unstable, most of the GAN-based models are trained with the addition of pixel losses~\cite{srgan,wang2018esrgan}.
To overcome the inherent problems of using pixel-based losses, Johnson et al.,~\cite{johnson2016perceptual} introduces the perceptual loss that calculates the distance between the embedded features of two output images.

To increase the perceptual quality, various studies have proposed unique network designs, losses, or network training techniques~\cite{blau20182018}.
In particular, EnhanceNet~\cite{enhancenet} and TSRN~\cite{TSRN} adopt texture matching loss~\cite{gatys2015texture} in combination with adversarial training and perceptual losses.
By providing texture information, the model can produce more realistic textures and reduce artifacts.
RankSRGAN~\cite{zhang2019ranksrgan} introduces ranker loss to reflect the perceptual metrics at the network optimization process.
ESRGAN~\cite{wang2018esrgan} improves the SRGAN by replacing the standard residual unit~\cite{mdsr2017} with the residual-in-residual dense block (RRDB) inspired by SRDenseNet~\cite{srdense}.
In addition, this model uses the relative discriminator loss~\cite{jolicoeur2018relativistic}.
However, the aforementioned models are not suitable for real-world applications despite the great visual quality of the SR output, because of the heavy computational requirements.
Although several studies have focused on designing a lightweight SR network~\cite{ignatov2018pirm,zhang2019aim}, the SR performance of the models is not satisfactory in terms of the perceptual quality.
On the contrary, our proposed models are able to generate visually plausible images with a reasonable amount of computation.

In the photo-realistic SR, measuring the quality of the resulting image is an open question.
The widely used distortion-based metrics such as PSNR and SSIM~\cite{ssim}, do not always reflect the human's perception of visual quality, and often contradict human judgment~\cite{srgan}.
To tackle this, various studies have proposed the perception-based image quality assessment in both full- and non-reference manners.
The full-reference metric measures the perceptual quality by using the distance between the features of the HR and generated images.
LPIPS~\cite{zhang2018perceptual} and NIQE~\cite{mittal2012making} are the notable assessments in this category.
On the other hand, the non-reference metric conducts the evaluation by predicting the human opinion score without ground-truth images.
NIMA~\cite{talebi2018nima}, Ma~\cite{ma2017learning} and PI~\cite{blau2018perception} fall into this approach.

\subsection{Efficient Neural Network}
\label{sec:efficient_nn}
There has been a rising interest in building a small and efficient network ~\cite{mobilenets,qin2020binary,gou2021knowledge}.
These approaches can be categorized into three groups: \textbf{1)} Compressing pretrained networks using pruning or quantizing techniques, \textbf{2)} transferring knowledge of a deep model to a shallow one, and \textbf{3)} designing small but efficient models.
In this section, we summarize the latter category, which aims to build a lean neural network in terms of design engineering, as it matches our approach most closely.

Iandola et al.,~\cite{squeezenet} introduces SqueezeNet to build a parameter-efficient architecture based on AlexNet~\cite{alexnet}.
By doing so, they achieve comparable classification accuracy with 50$\times$ fewer parameters than the baseline model.
Unlike SqueezeNet, MobileNet~\cite{mobilenets} aims to decrease the number of operations in order to reduce the inference runtime.
This model decomposes the standard convolution to 1$\times$1 and depthwise separable convolutions. 
While the MobileNet effectively cuts down the computational cost, 1$\times$1 convolution becomes the new bottleneck and thus can be the limitation to pushing down the overall cost.
To mitigate this issue, ShuffleNet variants~\cite{shufflenetv2} use the channel shuffle unit following the 1$\times$1 group convolution.
Referring to the recent literature~\cite{mobilenets,shufflenetv2}, we apply a depthwise separable convolution technique in residual blocks to build a fast and lightweight SR model.
Instead of using depthwise separable convolution, however, we use group convolution to make the efficiency of the network tunable.

\section{Our Methods}
\label{sec:method}
We design photo-realistic CARN (PCARN) by using our prior model, CARN~\cite{ahn2018fast}, as a generator.
We first recap our prior work, explain the modifications toward a better generator (\Sref{subsec:CARN}), and then introduce our PCARN.

\subsection{Cascading Residual Network}
\label{subsec:CARN}
\begin{figure}[t]
\centering
\begin{subfigure}[b]{0.45\linewidth}
	\includegraphics[width=\linewidth]{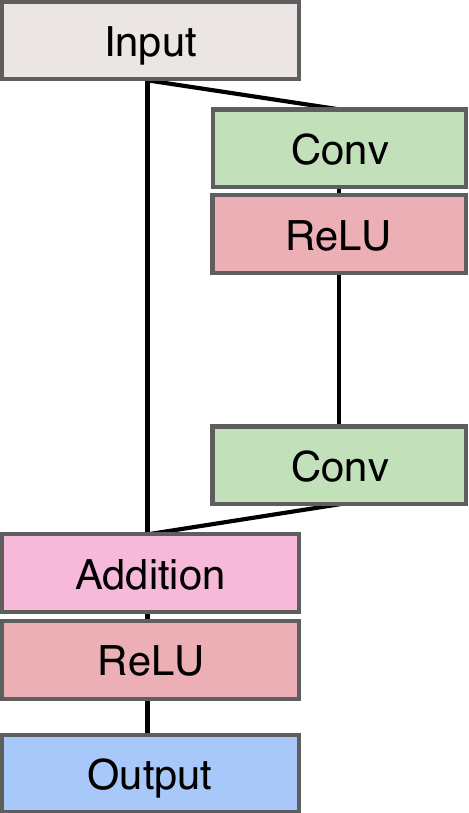}
	\caption{Residual Block}
\end{subfigure}
\begin{subfigure}[b]{0.45\linewidth}
	\includegraphics[width=\linewidth]{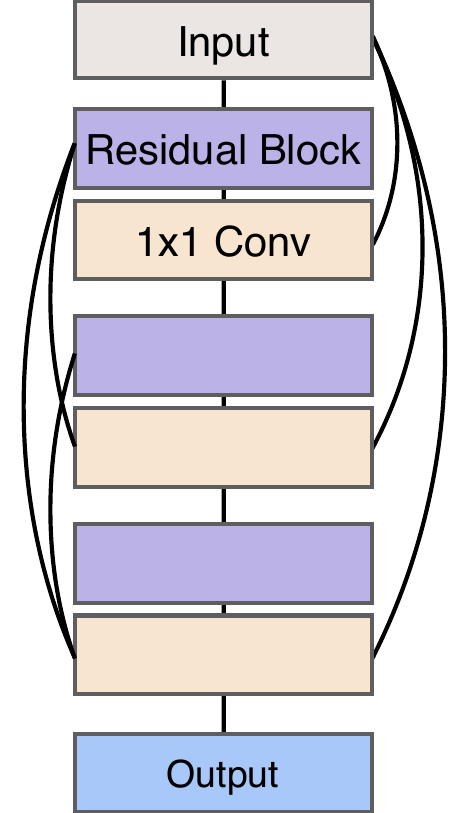}
	\caption{Cascading Block}
\end{subfigure}
\caption{\textbf{Structures of local blocks.} Our cascading block composed of residual blocks and local cascading connections.}
\label{fig:block}
\end{figure}

The main architecture of our generator (CARN) is based on the EDSR~\cite{mdsr2017}.
The prime difference between EDSR-like networks and ours is the presence of local and global cascading modules.
\Fref{fig:model} (top) graphically depicts how global cascading occurs.
The outputs of intermediary features are cascaded into the higher blocks and finally converge on a 1$\times$1 Conv layer.
Note that the intermediary modules are implemented as cascading blocks, which also host cascading connections themselves in a local way.
Such local cascading operations (\Fref{fig:block}b) is identical to a global one, except that the backbone units are the residual blocks.

To express how cascading works formally, we first define the standard residual block (\Fref{fig:block}a) as $R_i(H^{i-1};W_R^i),$ where $H^{i-1}$ is the input feature of the $i$-th residual block and $W_R^i$ is the parameter set of the convolution layers inside of each residual blocks.
Then, we replace the residual block with the local cascading block (\Fref{fig:block}b).
To formulate the local cascading as well, we denote $B^{i, j}$ as the output of the $j$-th residual block in the $i$-th cascading block, and $W_{local}^i$ as the set of parameters of the $i$-th local cascading block.
The $i$-th local cascading block $B_{local}^{i}$ is defined below.

\begin{equation}\label{eq:local_carn}
B_{local}^i\left(H^{i-1};W_{local}^i\right) \equiv B^{i, U},
\end{equation}
where $B^{i,U}$ is defined recursively from the $B^{i,u}$'s as:
\begin{align}
B^{i,0} &= H^{i-1} \notag \\
B^{i,u} &= g\left(\left[B^{i,0},\dots,B^{i,u-1},R^{u}\left(B^{i,u-1};W_R^{u}\right)\right]\right) \notag \\ 
& \quad \text{for $u=1,\dots,U$,}
\end{align}
where, $g$ is a 1$\times$1 convolution layer.
Finally, we define the output of the final cascading block $H^B$ by combining all $H^i$'s for $i=0,\cdots,b-1$.

\begin{align}\label{eq:carn}
H^0 &= f\left(\boldsymbol{X};W_c\right) \notag\\
H^B &= g\left(\left[H^0,\dots,H^{b-1},B^B_{local}\right(H^{b-1};W_B^b\left)\right]\right) \notag \\ 
& \qquad \text{for $b=1,\dots,B$,}
\end{align}
where $\mathbf{X}$ is the input LR image, $f$ is the first convolution layer (with parameter $W_c$) of the network, and $W^b_B$ is the parameter set of each cascading block.

On top of our preliminary work~\cite{ahn2018fast}, we make the following modifications to the model to boost up the SR performance.
\textbf{1)} Inspired by the VDSR~\cite{vdsr2016}, we adopt the global residual learning to our framework.
To do that, we aggregate the output of the entry layers and the final 1$\times$1 convolution layer right before the upsampling block.
Formally, it can be written as $O = H^b + H^0$, where the final feature map $O$ becomes the input to the upsampling block.
The effect of this final addition might appear redundant since the output of the first convolution is already added to the 1$\times$1 before being added again in the next step.
Nonetheless, we found that this duplicate addition is beneficial to the overall SR performance with little computational overhead.
\textbf{2)} We adjust the positions of ReLU in the network.
That is, we eliminate the nonlinearities following the 1$\times$1 convolution layer.
Additionally, we add nonlinearities in the upsampling unit to increase the expressive power of the network.

By applying the cascading mechanism on the local and global levels, we can get two advantages:
\textbf{1)} The model incorporates features from multiple layers, which allows learning multi-level feature representations.
\textbf{2)} The multi-level cascading connection operates as a multi-level shortcut connection that easily propagates information from lower to higher layers (and vice-versa, in the case of back-propagation).
Hence, the network can reconstruct the LR image based on multi-level features, and the upsampling unit also upsamples images by taking diverse features (from multiple layers) into account.
Thus, our design helps the model to boost the SR performance. We will show how such modules effectively work in \Sref{subsec:model_analysis}.

Inspired by VDSR~\cite{vdsr2016} and EDSR~\cite{mdsr2017}, we apply the multi-scale learning by embedding all up-sample blocks to a single network (\Fref{fig:model}).
The benefit of using such a strategy is that it can process multiple scales using a single trained model.
It also helps us alleviate the burden of multiple model size when deploying the SR application on small devices since our PCARN family only needs a single network for multiple scales.

\begin{figure}[tp]
\centering
\includegraphics[width=0.95\linewidth]{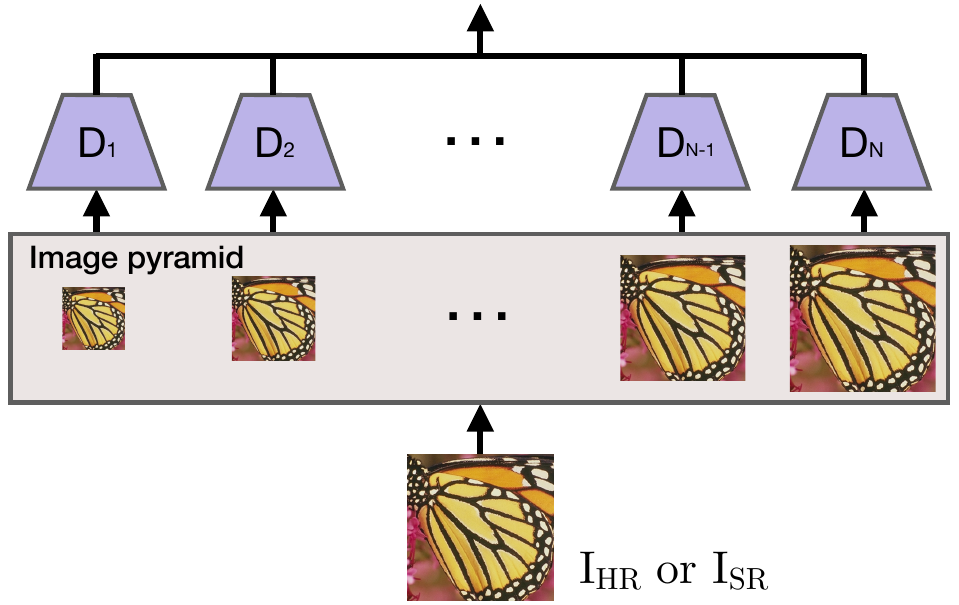}
\caption{\textbf{Conceptual illustration of the multi-scale discriminator.} Input image is downsampled using average pooling and each image is taken by the corresponding scale discriminator. Total discriminative loss is calculated by summing over all the scale losses.}
\label{fig:block_msd}
\end{figure}

\subsection{Photo-realistic CARN}
\label{subsec:PCARN}
Following \textit{Goodfellow et al.}~\cite{goodfellow2014generative}, we define a discriminator network $D$, which we optimize in an alternative procedure along with the generator $G$.
Using the discriminator and the generator, we denote the adversarial loss~\cite{goodfellow2014generative} as:

\begin{equation}\label{eq:gan_loss}
\begin{split}
L_{GAN}(G, D) &= \mathbb{E}_{I_{HR}}\left[\log D(I_{HR})\right] + \\
&\phantom{{}={}} \mathbb{E}_{I_{LR}}\left[\log(1-D(G(I_{LR})))\right],
\end{split}
\end{equation}
where $I_{HR}$ and $I_{LR}$ denote the HR and LR images, respectively.
The idea of adversarial loss is that it trains the generative model $G$ to fool the discriminator $D$, whereas the discriminator is trained to distinguish whether the images are from the SR or the HR sets.
This formulation encourages the generator to create perceptually superior images compared to the pixel-based (distortion-based) losses.

Many previous works have mixed the adversarial loss with a traditional pixel-based loss to stabilize the training process~\cite{srgan,wang2018esrgan}.
In this case, the task of a generator is not only to fool the discriminator but also to create an SR image similar to the HR.
We also take this option but use the VGG loss~\cite{johnson2016perceptual} instead of the pixel-based loss to avoid blurriness.
The VGG loss is defined as the distance between the outputs of the ReLU layers of the pre-trained VGG-19 network~\cite{vggnet2014}.
Formally, we denote the output feature map of the $j$-th ReLU following a convolutional layer before the $i$-th pooling layer as $\phi_{i,j}$.
Then, we define the VGG loss as the L2 distance between the feature representation of the HR image $I_{HR}$, and the super-resoluted image $G(I_{LR})$:

\begin{equation}\label{eq:vgg_loss}
\begin{split}
L_{VGG}(G) = \frac{1}{W_{i,j}H_{i,j}}\sum^{W_{i,j}}_{x}\sum^{H_{i,j}}_{y}\left[\phi_{i,j}(I_{HR})_{x,y}\right. \\
\left. \qquad -\phi_{i,j}(G(I_{LR}))_{x,y}\right]^2.
\end{split}
\end{equation}

Here, $W_{i,j}$ and $H_{i,j}$ are the spatial resolutions of the feature map. In our work, we use $i=j=5$.

To enhance the finer details of the outputs, we adopt the multi-scale discriminator (\Fref{fig:block_msd}).
The idea is to use multiple discriminators instead of a single one to make each discriminator handle a specific scale.
Thus, it allows the model to gather information across coarse- to fine-resolution images.
To do so, we downsample the input image (SR or HR) to make an image pyramid.
Then, the scaled images are fed into the corresponding discriminators and finally the multi-scale discriminator loss $L^{MS}_D$ is calculated by collecting each of the losses as in the equation below.

\begin{equation}
L^{MS}_D = \sum^{S}_{i} D_i(F_i(\mathbf{I})),
\end{equation}
where $\mathbf{I}$ is the input image and $F(.)$ is the scale-specific downsample function.
In all our experiments, we use average pooling as the downsampling module and set $S$ as three.

The total loss for the generator is computed by summing the multi-scale GAN and VGG losses as:
\begin{equation}
L_G = L^{MS}_{GAN} + \lambda L_{VGG},
\end{equation}
where $L^{MS}_{GAN}$ denotes the adversarial loss in terms of the generator with multi-scale discriminator and $\lambda$ is the hyperparameter to balance the two losses.

\subsection{Efficient Photo-realistic CARN}
\label{subsec:efficient_PCARN}
\begin{figure}[htp]
\centering
\begin{subfigure}[b]{0.45\linewidth}
	\includegraphics[width=\linewidth]{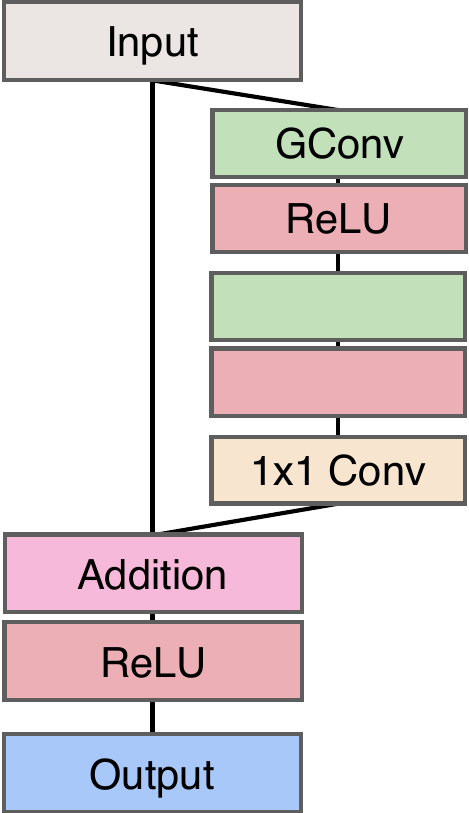}
	\caption{EResidual Block}
\end{subfigure}
\begin{subfigure}[b]{0.45\linewidth}
	\includegraphics[width=\linewidth]{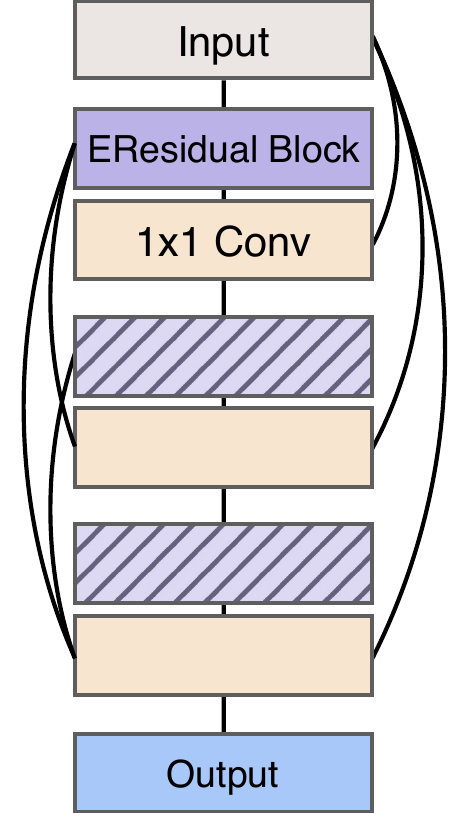}
	\caption{ECascading Block}
\end{subfigure}
\caption{\textbf{Structures of efficient cascading blocks.} (a) Efficient residual block, and (b) is the efficient cascading block. Hatched boxes in (b) denote the residual block with a parameter tying.}
\label{fig:block_efficient}
\end{figure}

To improve the efficiency of PCARN, we propose an efficient residual and cascading block of the generator.
This approach is analogous to the MobileNet~\cite{mobilenets}, but we use group convolution instead of depthwise separable convolution.
Our efficient residual (EResidual) block is composed of two consecutive 3$\times$3 group convolutions and a single pointwise convolution (\Fref{fig:block_efficient}a).
The advantage of using group convolution over the depthwise separable convolution is that it makes the efficiency of the model manually tunable.
Thus, the user can choose the appropriate group count for the desired performance, since the number of groups and the performance are in a trade-off relationship.

The analysis of the efficiency of the EResidual block usage is as follows.
Let $K$ be the kernel size and $C_{in}, C_{out}$ be the number of input and output channels.
Since we retain the spatial resolution of the feature map by the padding, we can denote $F$ to be both the input and output feature size.
Then, the cost of a standard residual block is

\begin{equation}
2\times\left(K^2 \cdot C_{in} \cdot C_{out} \cdot F^2\right).
\end{equation}

Note that we exclude the cost of addition or nonlinearity, and consider only the convolution layers.
This is because both the standard and the efficient blocks have the same number of such modules and these occupy a negligible portion of the entire computational cost.

Let $G$ be the number of groups. Then, the cost of an EResidual block, which consists of two group convolutions and one 1$\times$1 convolution, is as given in Equation~\ref{eq:reseblock}.

\begin{equation} \label{eq:reseblock}
2\times\left(K^2 \cdot C_{in} \cdot \frac{C_{out}}{G} \cdot F^2\right) + C_{in} \cdot C_{out} \cdot F^2
\end{equation}

By changing a standard residual block to our efficient block, we can reduce the computation by the ratio of

\begin{align}
& \frac{2\times\left(K^2 \cdot C_{in} \cdot \frac{C_{out}}{G} \cdot F^2\right) + C_{in} \cdot C_{out} \cdot F^2}{2\times\left(K^2 \cdot C_{in} \cdot C_{out} \cdot F^2\right)} \nonumber \\
&= \frac{1}{G} + \frac{1}{2K^2}.
\end{align}

Because we use a kernel size as 3$\times$3 for all convolutional layers, and the number of the channels is constant except the entry, exit, and upsampling block, the EResidual block reduces the computation from 1.8 up to 14 times depending on the number of groups.
To find the best trade-off between SR quality and computation cost, we perform an extensive case study (\Sref{subsec:efficiency}).

To further reduce the parameters, we apply a recursive network strategy.
In other words, we force the EResidual blocks to be shared in the cascading block, so only one-third of the parameters are needed compared to the standard block.
\Fref{fig:block_efficient}b shows our efficient cascading (ECascading) block after applying such scheme.
Unlike the previous studies that adopt the recursive scheme~\cite{kim2016deeply,MSLapSRN}, we do not increase the depth or the width of the network, so the number of operations is kept the same.

\subsection{Comparison to Recent Models}
\noindent\textbf{Comparison to MemNet.}
MemNet~\cite{memnet} and ours have a similar motivation, but there are two main distinctions from our schemes.
\textbf{1)} Feature fusion is done in a different location and manner.
For instance, MemNet fuses the output features of each recursive unit at the end of the memory blocks.
On the other hand, we gather the information at every possible site in the local block, thus can boost up the representation power via additional layers.
\textbf{2)} MemNet takes an \textit{early-upsample} approach which upsamples the image before giving it to the model.
Although it becomes easier to implement residual learning, it worsens the model efficiency substantially.
In contrast, our model gets LR images and intermediate features are upsampled at the end of the network, which enables us to accomplish a good balance between the SR quality and efficiency.

\noindent\textbf{Comparison to DenseNet.}
SRDenseNet~\cite{srdense} uses a densely connected block and skip connection.
Although the overall design concept can be similar, our model has two main advantages. 
\textbf{1)} In our models, the output of each block is associated with a global cascading connection which is a generalized form of the skip connection.
In SRDenseNet, all levels of features are combined after the final dense block, but our global cascading scheme connects all blocks, which behaves as a multi-level skip connection.
\textbf{2)} The connectivity schemes that we use are economical for both memory and speed.
In a densely connected block~\cite{srdense}, concatenated features are distilled to reduce the number of channels only at the end of the block.
Such a block design can allow fluent information flow (since no channel reducing operation exists), but it requires a high amount of computation because blocks have to carry all the intermediate features.
In contrast, we incorporate features using an additional 1$\times$1 convolution layer at each concatenation point, which facilitates composing more lightweight models.

\subsection{Implementation Details}
For the PCARN generator, we set $B=U=3$ and the number of channels in all convolutional layers, except for the first, last layer and upsample block, to 64.
For the upsampling unit, we use the \textit{pixelshuffle} operation~\cite{espcn2016} following the convolutional layer.
Our discriminator network has 9 convolutional layers (\Fref{fig:model}).
We train our models with ADAM by setting $\beta_{1}=0.9$, $\beta_{2}=0.999$, and $\epsilon = 10^{-8}$ in $6 \times 10^5$ steps.
The minibatch and patch sizes are 64 and 48$\times$48, respectively.
We use initial learning rate as $10^{-4}$ and halved every $4 \times 10^5$ steps.

Following SRGAN~\cite{srgan}, we first train the generator with L1 loss, then fine-tune the pretrained network for $2\times10^5$ steps with the same settings but using GAN loss.
When training the generator in a pixel-wise manner, we use the L1 loss as our loss function instead of the L2.
The L2 loss is widely used in the image restoration task because of its relationship to the PSNR, but L1 provided better convergence and performance in our experiments.

We use DIV2K~\cite{div2k}, which consists of 800 training and 100 validation images.
Because of the richness of this dataset, recent SR models~\cite{mdsr2017,wang2018esrgan} use DIV2K as well.
To prepare the training input, we randomly crop images to the 48$\times$48 LR patches and augment to horizontal flip or rotation.
To enable the multi-scale training, we first randomly select the scale from [2, 4]. Then we construct the training batch using chosen scale, since our model can process only a single scale for each batch.
For the test and benchmark, Set5~\cite{set5}, Set14~\cite{yang2010}, B100~\cite{b100} and Urban100~\cite{urban100} datasets are used.
The code is publicly available\footnote{\url{https://github.com/nmhkahn/PCARN-pytorch}}.

\begin{figure}[t]
\centering
\includegraphics[width=0.92\linewidth]{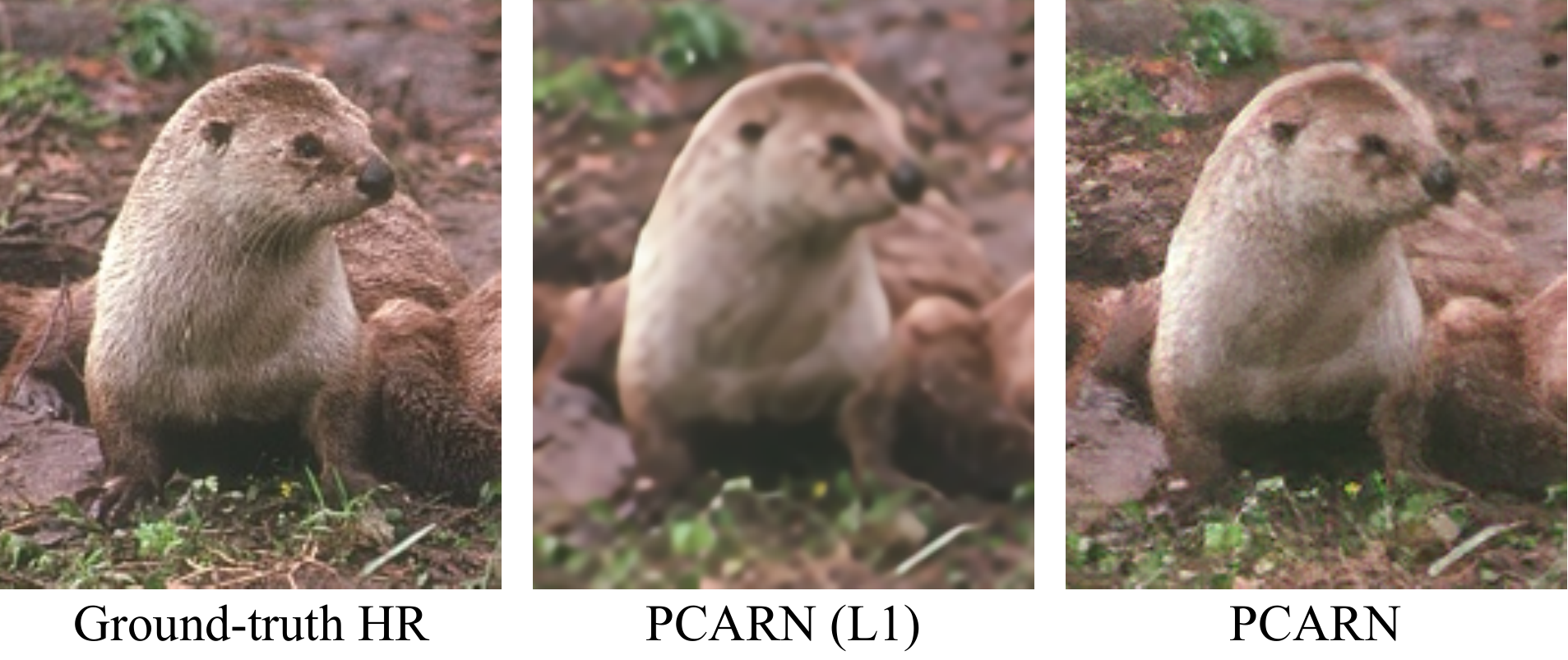}
\caption{\textbf{Visual comparison of adversarial training.}  We compare the results trained with PCARN and without the adversarial training, PCARN (L1).}
\label{fig:comp_gan}
\end{figure}

\section{Model Analysis}
In this section, we first present the internal analysis of our model.
Here, we train and evaluate each of the methods ten times and gather the mean and standard deviation to inspect the performance more accurately.
One thing to note here is that we represent the number of operations by MultAdds. It is the number of composite multiply-accumulate operations for a single image. We assume the HR image to be 720p (1280$\times$720) to calculate MultAdds.
For the internal performance evaluation on the quantitative view, we use LPIPS~\cite{zhang2018perceptual} and NIMA~\cite{talebi2018nima} scores to capture the perceptual quality of the output images.
We use the LPIPS settings of the AlexNet~\cite{alexnet} with the fine-tuned linear layer (AlexNet-linear, version 0.1), and MobileNet~\cite{mobilenets} for the NIMA backbone model.

\subsection{Model Design}
\label{subsec:model_analysis}
To investigate the performance of the proposed methods, we analyze our models via ablation study.
We select the baseline to be SRResNet~\cite{srgan}. Other models have the same topology except for the inherent modules (\textit{e.g.,} additional 1$\times$1 convolution) that are needed for each particular architecture.
Thus, the overall number of parameters is increased by up to 15\% from the baseline.

\begin{table}[t]
\setlength{\tabcolsep}{9.3pt}
\begin{center}
\begin{tabular}{l r c c}
\hline\Tstrut
\textbf{Model} & \textbf{Params} & \textbf{PSNR} & \textbf{SSIM} \Bstrut\\\hline\Tstrut
Baseline       & 963K   & 28.42\scriptsize{$\pm$0.02}	        & 0.7773\scriptsize{$\pm$3e-4} \\
+ Local        & 1,074K	& 28.45\scriptsize{$\pm$0.01}	        & 0.7780\scriptsize{$\pm$3e-4} \\
+ Global       & 1,000K	& 28.47\scriptsize{$\pm$0.02}	        & 0.7787\scriptsize{$\pm$4e-4} \\
+ L/G           & 1,111K	& 28.49\scriptsize{$\pm$0.02}	        & 0.7788\scriptsize{$\pm$3e-4} \\
+ Residual     & 1,111K	& \textbf{28.50\scriptsize{$\pm$0.01}}	& \textbf{0.7792\scriptsize{$\pm$3e-4}}\Bstrut\\\hline
\end{tabular}
\end{center}
\caption{\textbf{Analysis of the effect of model design choices.} \textit{Local}, \textit{Global} and \textit{L/G} means the models with local, global, and both cascading connection respectively.}
\label{table:ablation}
\end{table}

\begin{table}[t]
\begin{center}
\begin{tabular}{l c c c}
\hline\Tstrut
\textbf{Model} & \textbf{LPIPS} & \textbf{PSNR} & \textbf{SSIM} \Bstrut\\\hline\Tstrut
PCARN (L1) & 0.289\scriptsize{$\pm$1e-3}          & 28.50\scriptsize{$\pm$0.01} & 0.7792\scriptsize{$\pm$3e-4} \\
+ GAN      & 0.162\scriptsize{$\pm$5e-3}          & 26.36\scriptsize{$\pm$0.25} & 0.7120\scriptsize{$\pm$6e-3} \\
+ MSD      & \textbf{0.155\scriptsize{$\pm$2e-3}} & 26.10\scriptsize{$\pm$0.17} & 0.6980\scriptsize{$\pm$8e-3}\Bstrut\\\hline
\end{tabular}
\end{center}
\caption{\textbf{Model analysis study.} MSD indicates PCARN model trained with multi-scale discriminator.}
\label{table:ablation_perception}
\end{table}

\begin{table}[t]
\begin{center}
\setlength{\tabcolsep}{8pt}
\begin{tabular}{c c c c}
\hline\Tstrut
\textbf{Dist.} & \textbf{std/range} & \textbf{PSNR} & \textbf{SSIM} \Bstrut\\\hline\Tstrut
Norm. & $0.1\times\sqrt{2/F}$ & 28.46\scriptsize{$\pm$0.01} & 0.7781\scriptsize{$\pm$3e-4} \\
Unif. & $0.1\times\sqrt{6/F}$ & 28.47\scriptsize{$\pm$0.01} & 0.7782\scriptsize{$\pm$3e-4} \\
Unif. & $0.1\times\sqrt{1/F}$ & 28.45\scriptsize{$\pm$0.02} & 0.7775\scriptsize{$\pm$4e-4} \\
Norm. & $1.0\times\sqrt{2/F}$ & 28.45\scriptsize{$\pm$0.02} & 0.7777\scriptsize{$\pm$3e-4} \\
Unif. & $1.0\times\sqrt{6/F}$ & 28.44\scriptsize{$\pm$0.01} & 0.7778\scriptsize{$\pm$2e-4} \\
Unif. & $1.0\times\sqrt{1/F}$ & \textbf{28.50\scriptsize{$\pm$}0.01} & \textbf{0.7791\scriptsize{$\pm$}3e-4}\Bstrut\\\hline
\end{tabular}
\end{center}
\caption{\textbf{Effect of the initialization.} $F$ denotes the number of the input channels (fan-in). \textbf{std/range} behaves as a standard deviation for the zero-mean normal distribution ($N(0, \text{std})$), or as a range for uniform distribution ($\text{Unif}(-\text{range}, \text{range})$).}
\label{table:init}
\end{table}

\Tref{table:ablation} shows the model analysis on the effect of cascading modules and the global residual learning scheme.
Model with local cascading improves the baseline SR performance.
We conjecture that this is because the cascading module passes not only the inputs but also the mixture of intermediate features to the next block, thus leveraging multi-level representations.
By incorporating multi-level representations, the model can consider a variety of information from many different receptive fields when reconstructing the image.
We observed higher performance gain with the global cascading scheme.
This is because the advantages of the local scheme are limited to each block, which lessens the model's ability to exploit the cascading effect.
One major benefit of the global cascading is that it allows information integration from lower layers, and this information shortcut provides useful clues for reconstructing the HR image in the upsampling and final reconstructing processes.
In addition, we employ the global residual learning shown in many recent SR methods~\cite{vdsr2016}.
The benefit of using it can be minor, since the roles of the global cascading and residual learning overlap.
However, we choose to embed the global residual learning to our framework since it does improve performance with negligible overhead.

\begin{figure}[t]
\centering
\begin{subfigure}[b]{\linewidth}
	\includegraphics[width=\linewidth]{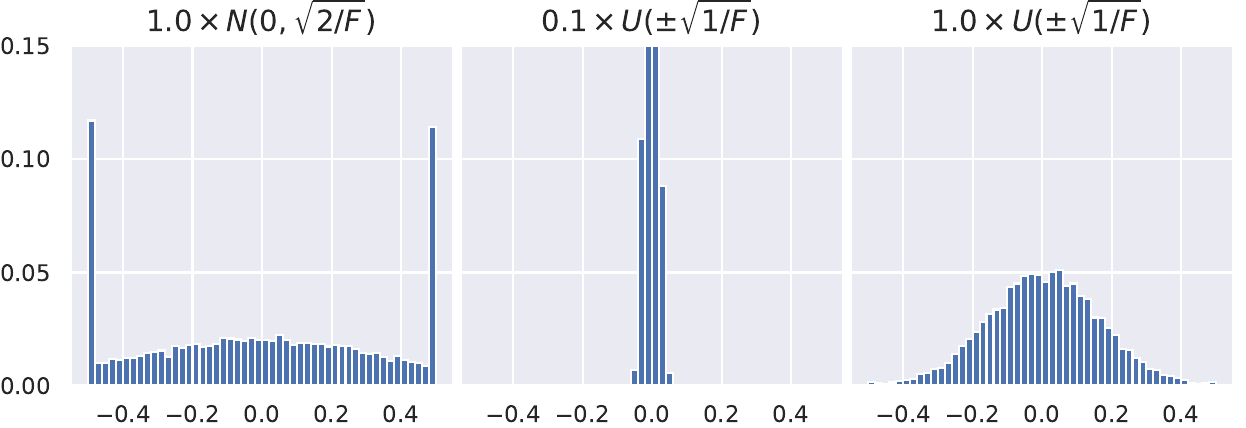}
	\caption{Activation values normalized histogram from the 1st layer.}
\end{subfigure}
\\
\begin{subfigure}[b]{\linewidth}
	\includegraphics[width=\linewidth]{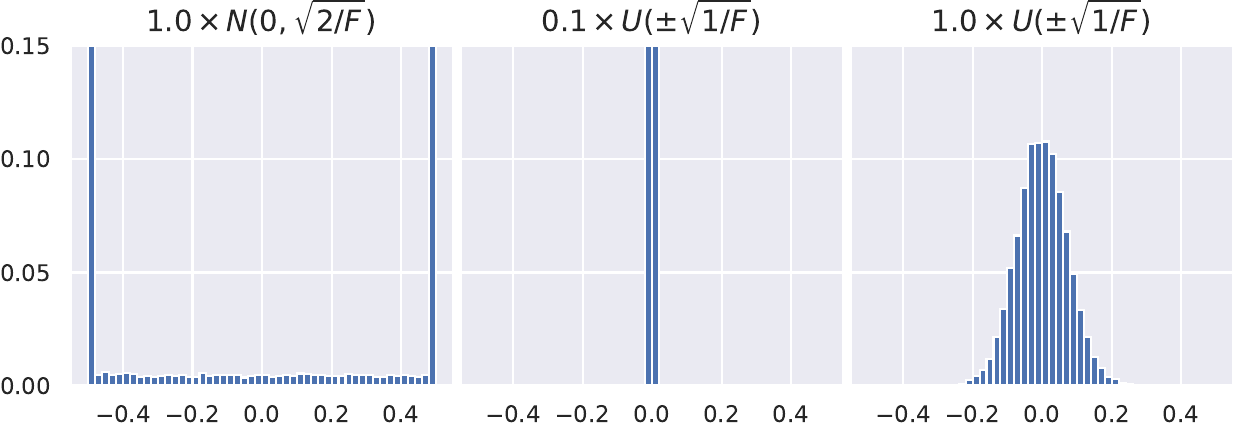}
	\caption{Activation values normalized histogram from the 2nd cascading block.}
\end{subfigure}
\\
\begin{subfigure}[b]{\linewidth}
	\includegraphics[width=\linewidth]{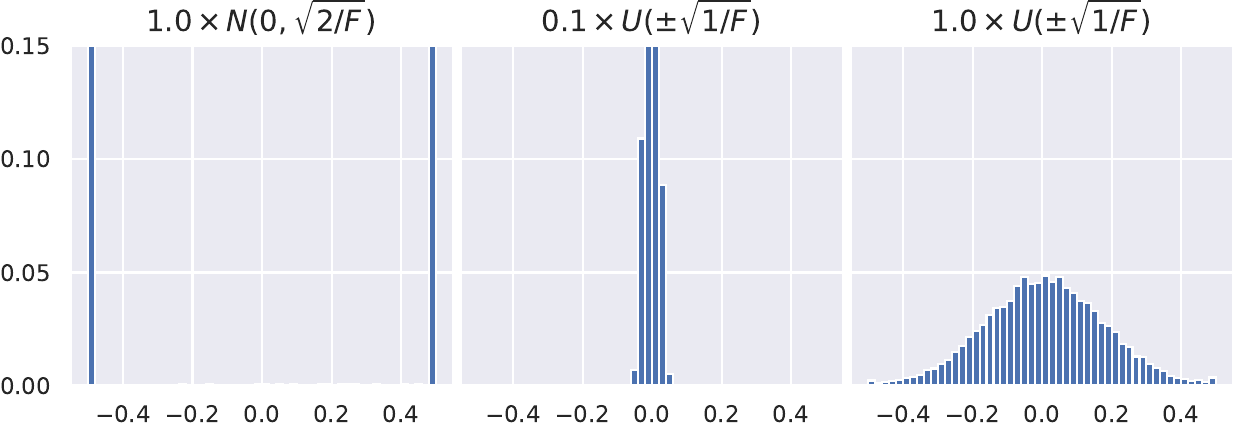}
	\caption{Activation values normalized histogram from the last cascading block.}
\end{subfigure}
\caption{\textbf{Activation values normalized histogram from different initializations.} We plot the activation (before ReLU) from the first layer (a), middle block (b), and final block (c). A model initialized with $1.0\times U(\pm\sqrt{1/F})$ (right) effectively carries signal to the last block, while others tend to saturate activations. Note that we clip the values to [-0.5, 0.5] and normalize the histogram.}
\vspace{-1em}
\label{fig:init_analysis}
\end{figure}

To build a photo-realistic SR model, it is essential to design a well-functioning discriminator.
To examine how the choice of discriminator affects the performance, we conducted a series of comparisons across various types of discriminator losses by using LPIPS~\cite{zhang2018perceptual}.
As shown in \Tref{table:ablation_perception}, PCARN with adversarial training (\textit{+GAN}) outperforms the baseline by a large margin.
\Fref{fig:comp_gan} also shows the advantage of using GAN, where it successfully recovers the fine details and generates more photo-realistic images.
However, because the model is not directly optimized using pixel-based loss, the performance of pixel-based metrics (PSNR and SSIM) is degraded.
Moreover, the overall training process is substantially unstable and shows high variance in all metrics.
Using the multi-scale discriminator (\textit{+MSD}) also gives an additional gain to LPIPS.
The main reason is that all the generators can distinguish between fake and real more easily since each generator covers different receptive fields.
This gives the generator more useful information to restore both fine- and coarse-level structures.

\subsection{Initialization Strategy}
\label{subsec:init}
Appropriate network initialization is the key component for boosting performance.
To verify the optimal one for our model, we performed an experiment comparing to six common initialization schemes: uniform, and normal distribution with various settings, as shown in \Tref{table:init}.
Note that we conduct this experiment using the PCARN with L1 loss since when training a model with GAN loss, we use the pre-trained network (with L1 loss) as the starting point.
Interestingly, the MSRA initialization~\cite{he2015delving} (4th row) and the high-range uniform (5th row) were inferior to the other methods.
We argue that a narrow 1$\times$1 convolution affects the quality of initialization since the high-variance initial values tend to result in high-variance activations.
Multiplying the initial random values by 0.1 (1st$\sim$3rd rows) degrades the performance as well.

\begin{figure}[t]
\centering
\includegraphics[width=0.97\linewidth]{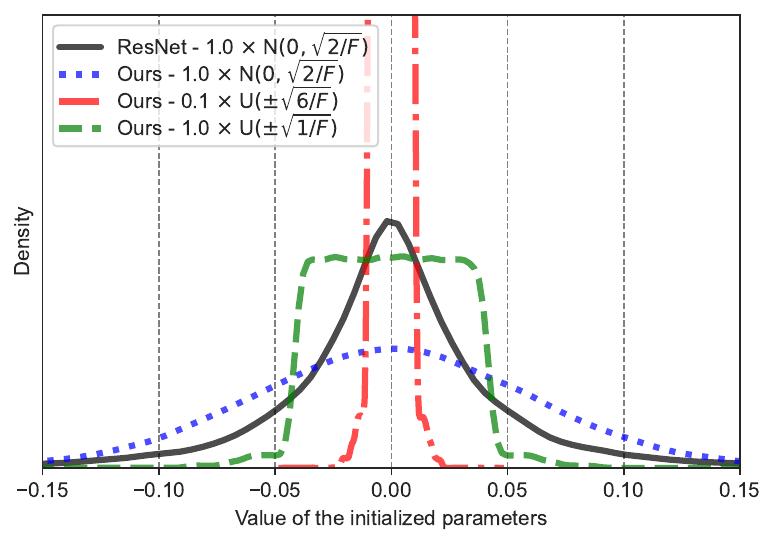}
\caption{\textbf{Distribution for the randomly initialized networks.} We plot 100K parameters for each model using 1000 bins. Each of the parameter distributions is from the diverse sources because of the variation in the number of input channels.}
\label{fig:init}
\end{figure}

We hypothesize that the degraded SR performance of other initialization schemes is mainly due to the saturated activations of the network.
If the activation generated from the deep layer saturates, useful information can be lost and the gradient signal can vanish, which results in poor performance of the model~\cite{glorot2010}.
As shown in \Fref{fig:init_analysis}, initializing with $1.0\times U(\pm\sqrt{1/F})$ does not suffer the saturation behavior, while others drive the activations toward zero or infinity.
\Fref{fig:init} explains why such initialization strategies suffer saturation.
Unlike the ResNet~\cite{he2016deep} results (black solid), the weights of our network initialized with $1.0\times N(\sqrt{2/F})$ (blue dots) have high-variance due to the narrow 1$\times$1 convolutions, so the output activations can be large.
On the other hand, initializing with $0.1\times U(\pm\sqrt{1/F})$ (red dots) makes the range of the parameters too narrow, thus saturating activations in deep layers toward zero.

\subsection{Efficiency Trade-off}
\label{subsec:efficiency}
\begin{figure}[tp]
\centering
\includegraphics[width=\linewidth]{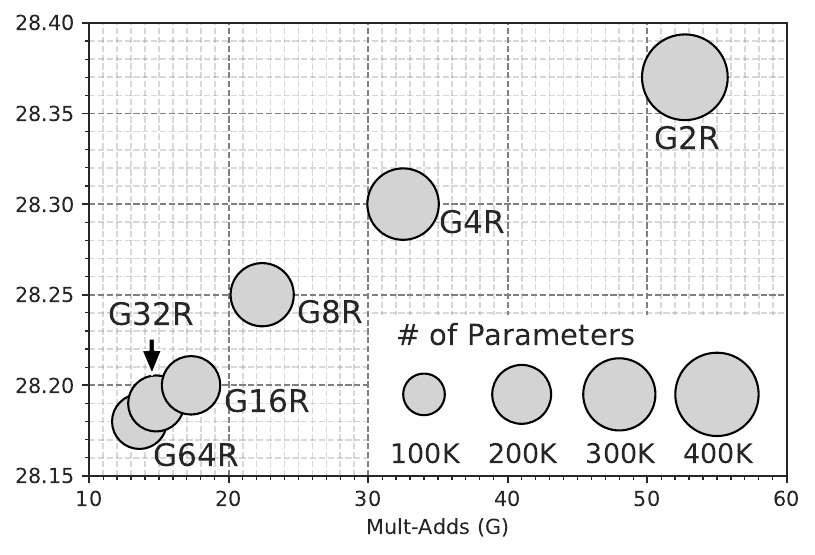}
\caption{\textbf{Efficiency analysis of efficient models.} We evaluate all models on Set14 with $\times$4 scale. \texttt{G} represents the number of groups of the group convolution, and \texttt{R} means the model with the recursive network scheme.}
\label{fig:mobile}
\end{figure}

\Fref{fig:mobile} depicts the trade-off analysis between the SR performance and efficiency of the efficient PCARN that uses convolution and a recursive scheme.
We use the model with L1 loss and evaluate using pixel-based metrics, PSNR and SSIM.
Although all efficient models perform worse than the PCARN, the number of parameters and operations are decreased dramatically.
We choose \texttt{G4R} as the best-balanced model, which we denote as PCARN-M, since the effect of compressing the model is reduced when the number of groups is larger than four.
As a result, PCARN-M reduces the number of parameters by four times and the number of operations by nearly three times with a 0.20 dB loss in PSNR and 0.0053 in SSIM, compared to the PCARN.

We observed that depthwise separable convolution (\texttt{G64R}) extremely degrades the performance (-0.32 dB).
There can be many explanations why such an observation occurs, but we suspect that this is because the image recognition and generation tasks are entirely different, so applying group and depthwise convolution, which are mainly used in recognition fields, has to be done very carefully.
Therefore, creating efficient SR model needs more investigation with plenty of room to improve performance.

\begin{table*}[htp]
\begin{center}
\bgroup
\def\arraystretch{0.95}
\setlength{\tabcolsep}{4.5pt}
\begin{tabular}{l r | c c c | c c c c | c c}
\hline\Tstrut
\multirow{2}{*}{Dataset} & \multirow{2}{*}{Metric} & \multicolumn{3}{c|}{\textbf{Distortion-based}} & \multicolumn{4}{c|}{\textbf{Perception-based}} & \multicolumn{2}{c}{\textbf{Ours}}
\\\cline{3-11}
& & SRCNN & MSLapSRN & SRResNet & FEQE & SRGAN & ENet & TSRN-G & PCARN & PCARN-M \\\hline\hline
\multirow{2}{*}{\textbf{-}} & Params & 0.1M & 0.2M & 1.5M & 0.1M & 1.5M & 1.1M & 1.1M & 1.6M & 0.4M \\
& MAs   & 52.7G & 435.9G & 127.8G & 5.6G & 127.8G & 120.6G & 120.6G & 90.9G & 32.5G \\\hline
\multirow{4}{*}{\textbf{Set5}}
& PSNR ↑ & 30.48 & \blue{31.74}  & \red{32.05} & 31.32 & 28.70 & 28.85 & 28.98 & 29.75 & 29.38 \\
& SSIM ↑ & 0.863 & \blue{0.889}  & \red{0.891} & 0.875 & 0.828 & 0.816 & 0.829 & 0.839 & 0.837 \\
& LPIPS ↓ & 0.201 & 0.178 & 0.173 & 0.142 & 0.088 & 0.099 & 0.088 & \red{0.075} & \blue{0.080} \\
& NIQE ↓ & 8.525 & 7.530  & 7.204 & 7.417 & 4.810 & \blue{4.411} & \red{4.205} & 4.586 & 4.904 \\
\hline\multirow{4}{*}{\textbf{Set14}}
& PSNR ↑ & 27.49 & \blue{28.26}  & \red{28.53} & 28.09 & 25.51 & 26.03 & 25.87 & 26.57 & 26.31 \\
& SSIM ↑ & 0.750 & \blue{0.774}  & \red{0.780} & 0.766 & 0.676 & 0.686 & 0.689 & 0.710 & 0.709 \\
& LPIPS ↓ & 0.315 & 0.299 & 0.284 & 0.256 & 0.174 & 0.162 & 0.155 & \red{0.145} & \blue{0.150} \\
& NIQE ↓ & 6.791 & 6.698  & 5.858 & 6.348 & 4.043 & 4.075 & \blue{3.633} & \red{3.501} & 3.727 \\
\hline\multirow{4}{*}{\textbf{B100}}
& PSNR ↑ & 26.90 & \blue{27.43}  & \red{27.57} & 27.23 & 24.37 & 25.26 & 25.19 & 25.56 & 25.48 \\
& SSIM ↑ & 0.710 & \blue{0.731}  & \red{0.735} & 0.723 & 0.619 & 0.635 & 0.649 & 0.653 & 0.657 \\
& LPIPS ↓ & 0.410 & 0.389 & 0.375 & 0.351 & 0.203 & 0.210 & \blue{0.196} & \red{0.188} & 0.198 \\
& NIQE ↓ & 6.876 & 6.699  & 6.188 & 6.726 & 4.054 & 4.506 & \blue{3.564} & \red{3.560} & 3.615 \\
\hline\multirow{4}{*}{\textbf{Urban}}
& PSNR ↑ & 24.52 & \blue{25.51}  & \red{26.07} & 25.26 & 23.75 & 23.62 & 23.68 & 24.17 & 23.73 \\
& SSIM ↑ & 0.722 & \blue{0.768}  & \red{0.784} & 0.755 & 0.706 & 0.693 & 0.705 & 0.719 & 0.704 \\
& LPIPS ↓ & 0.316 & 0.252 & 0.226 & 0.241 & 0.156 & 0.171 & \blue{0.154} & \red{0.152} & 0.167 \\
& NIQE ↓ & 6.359 & 5.777  & 5.311 & 6.185 & 3.829 & 3.716 & \red{3.561} & 3.598 & \blue{3.566} \\

\hline
\end{tabular}
\egroup
\end{center}
\caption{\textbf{Comparison for $\times$4 SISR on full-reference quality measures.} MAs denotes MultAdds.}
\label{table:benchmark_fr}
\end{table*}

\begin{table*}[htp]
\begin{center}
\bgroup
\def\arraystretch{0.95}
\setlength{\tabcolsep}{4.5pt}
\begin{tabular}{l r | c c c | c c c c | c c}
\hline\Tstrut
\multirow{2}{*}{Dataset} & \multirow{2}{*}{Metric} & \multicolumn{3}{c|}{\textbf{Distortion-based}} & \multicolumn{4}{c|}{\textbf{Perception-based}} & \multicolumn{2}{c}{\textbf{Ours}}
\\\cline{3-11}
& & SRCNN & MSLapSRN & SRResNet & FEQE & SRGAN & ENet & TSRN-G & PCARN & PCARN-M \\\hline\hline
\multirow{2}{*}{\textbf{-}} & Params & 0.1M & 0.2M & 1.5M & 0.1M & 1.5M & 1.1M & 1.1M & 1.6M & 0.4M \\
& MAs   & 52.7G & 435.9G & 127.8G & 5.6G & 127.8G & 120.6G & 120.6G & 90.9G & 32.5G \\\hline
\multirow{3}{*}{\textbf{Set5}}
& Ma ↑	& 4.865 & 5.315	        & 5.538	        & 5.563	        & \blue{7.965}	        & \red{8.111}	        & 7.900       	& 7.794       	& 7.810 \\
& PI ↓	& 6.830 & 6.108	        & 5.833	        & 5.927	        & 3.423	        & \red{3.150}           & \blue{3.153}       	& 3.396       	& 3.547 \\
& NIMA ↑& 4.316 & 4.669 & 4.743 & 4.480 & \blue{4.865} & 4.859 & 4.798 & \red{4.929} & \blue{4.865} \\
\hline\multirow{3}{*}{\textbf{Set14}}
& Ma ↑	& 4.584 & 5.359	        & 5.799	& 5.551	                 & \red{8.095}	        & 8.046	        & \blue{8.064}       	& 7.990       	& 7.793 \\
& PI ↓	& 6.104 & 5.669	        & 5.030	& 5.399	                & 2.974	        & 3.014	        & \blue{2.784}       	& \red{2.755}       	& 2.967 \\
& NIMA ↑& 4.372 & 4.797 & 4.910 & 4.594 & 5.002 & \red{5.127} & \blue{5.071} & 5.063 & 4.981 \\
\hline\multirow{3}{*}{\textbf{B100}}
& Ma ↑	& 4.742 & 5.476	        & 5.790	& 5.453	              & \red{8.741}	        & 8.549	        & \blue{8.655}       	& 8.610       	& 8.569 \\
& PI ↓	& 6.067 & 5.612	        & 5.199	& 5.636	              & 2.657	        & 2.978	        & \red{2.454}       	& \blue{2.475}       	& 2.523 \\
& NIMA ↑& 4.340 & 4.609 & 4.685 & 4.427 & 4.936 & \blue{5.067} & 4.841 & \red{5.077} & 5.010 \\
\hline\multirow{3}{*}{\textbf{Urban}}
& Ma ↑	& 4.442 & 5.152	        & 5.529	& 5.191	              & \blue{6.896}	        & \red{6.915}	        & 6.755       	& 6.820       	& 6.818 \\
& PI ↓	& 5.958 & 5.312	        & 4.891	& 5.497	              & 3.466	        & 3.400	        & 3.403	        & \blue{3.389}       	& \red{3.374} \\
& NIMA ↑& 4.624 & 5.039 & 5.135 & 4.879 & 5.191 & \blue{5.201} & 5.143 & \red{5.243} & 5.180 \\
\hline
\end{tabular}
\egroup
\end{center}
\caption{\textbf{Comparison for $\times$4 SISR on non-reference quality measures.} MAs denotes MultAdds.}
\label{table:benchmark_nr}
\end{table*}

\begin{table*}[htp]
\begin{center}
\bgroup
\def\arraystretch{0.95}
\setlength{\tabcolsep}{10pt}
\begin{tabular}{l r | c c | c c | c c}
\hline\Tstrut
\multirow{2}{*}{Dataset} & \multirow{2}{*}{Metric} & \multicolumn{2}{c|}{\textbf{Scale $\times2$}} & \multicolumn{2}{c|}{\textbf{Scale $\times3$}} & \multicolumn{2}{c}{\textbf{Scale $\times4$}} \\\cline{3-8}
& & PCARN & PCARN-M & PCARN & PCARN-M & PCARN & PCARN-M \\\hline\hline
\multirow{5}{*}{\textbf{Set5}}
& LPIPS ↓ & 0.019 & 0.023 & 0.044 & 0.053 & 0.075 & 0.080 \\
& NIQE ↓ & 4.765 & 4.359 & 4.905 & 4.908 & 4.586 & 4.904 \\
& Ma ↑   & 8.146 & 8.160 & 7.889 & 7.720 & 7.794 & 7.810 \\
& PI ↓ & 3.309 & 3.100 & 3.508 & 3.594 & 3.396 & 3.547 \\
& NIMA ↑  & 4.872 & 4.845 & 4.858 & 4.760 & 4.929 & 4.865  \\
\hline
\multirow{5}{*}{\textbf{Set14}}
& LPIPS ↓ & 0.045 & 0.049 & 0.096 & 0.106 & 0.145 & 0.150 \\
& NIQE ↓ & 3.857 & 3.841 & 3.644 & 3.737 & 3.501 & 3.727 \\
& Ma ↑  & 8.059 & 8.045 & 8.065 & 7.918 & 7.990 & 7.793 \\
& PI ↓  & 2.899 & 2.898 & 2.789 & 2.910 & 2.755 & 2.967 \\
& NIMA ↑ & 5.118 & 5.044 & 5.132 & 5.009 & 5.063 & 4.981 \\
\hline
\multirow{5}{*}{\textbf{B100}}
& LPIPS ↓ & 0.060 & 0.064 & 0.129 & 0.139 & 0.188 & 0.198 \\
& NIQE ↓ & 3.814 & 3.666 & 3.662 & 3.502 & 3.560 & 3.615 \\
& Ma ↑  & 8.748 & 8.756 & 8.673 & 8.624 & 8.610 & 8.569 \\
& PI ↓  & 2.533 & 2.455 & 2.495 & 2.439 & 2.475 & 2.523 \\
& NIMA ↑ & 4.980 & 4.936 & 5.074 & 4.985 & 5.077 & 5.010 \\
\hline
\multirow{5}{*}{\textbf{Urban}}
& LPIPS ↓ & 0.040 & 0.047 & 0.095 & 0.108 & 0.152 & 0.167 \\
& NIQE ↓ & 3.974 & 4.170 & 3.860 & 3.888 & 3.598 & 3.566 \\
& Ma ↑ & 6.751 & 6.734 & 6.769 & 6.724 & 6.820 & 6.818 \\
& PI ↓  & 3.612 & 3.718 & 3.545 & 3.582 & 3.389 & 3.374 \\
& NIMA ↑ & 5.240 & 5.204 & 5.214 & 5.177 & 5.243 & 5.180 \\\hline
\end{tabular}
\egroup
\end{center}
\caption{\textbf{Records of our models in diverse scales on perception-based image quality assessments.}}
\label{table:ours_perception}
\end{table*}

\begin{figure*}[tp]
\centering
\includegraphics[width=0.9\linewidth]{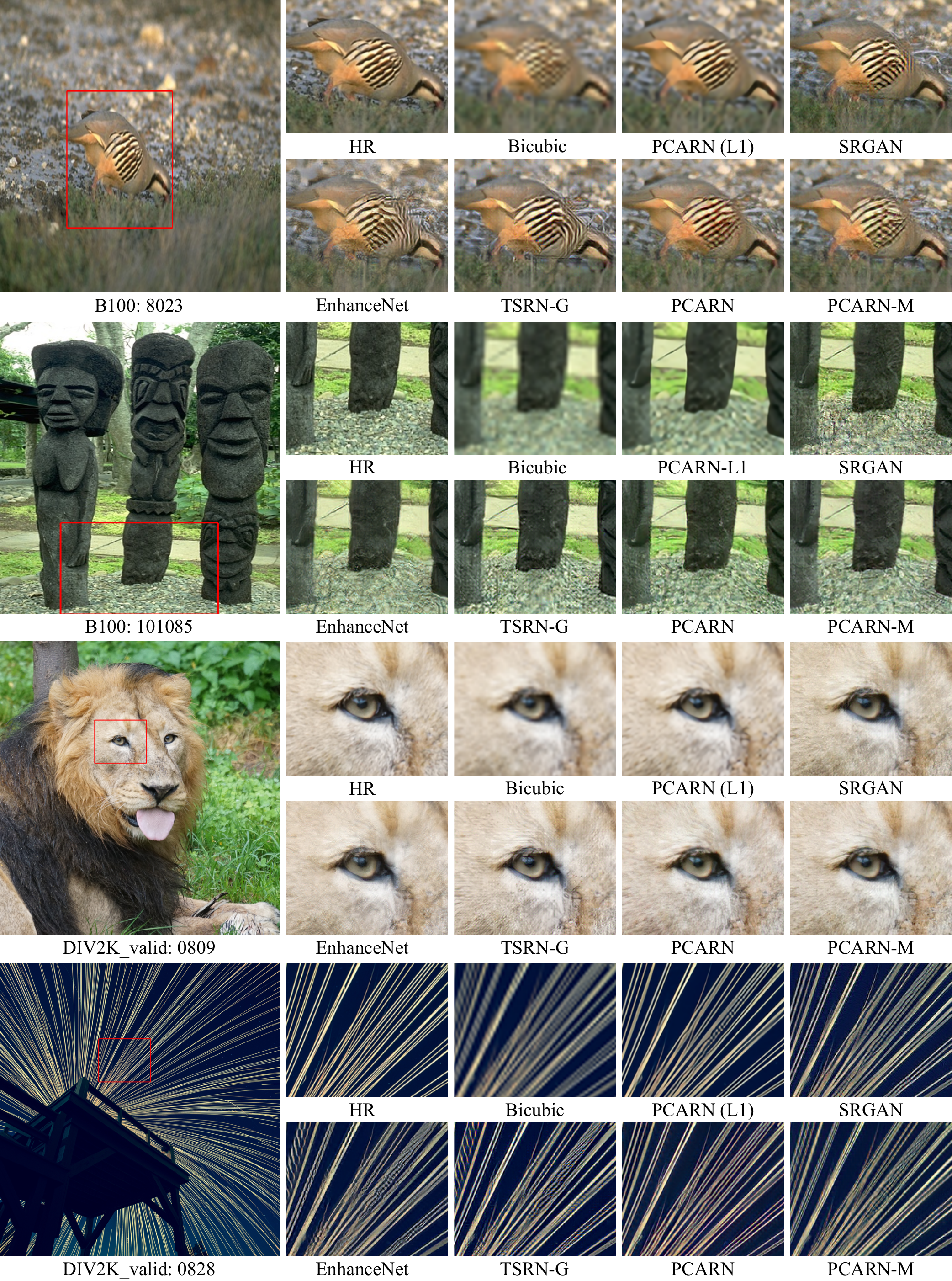}
\caption{\textbf{Visual comparison.} We compare perception-based methods that have similar complexity to ours on the $\times$4 scale SR datasets.}
\label{fig:comp_perception}
\end{figure*}

\begin{figure*}[tp]
\centering
\includegraphics[width=\linewidth]{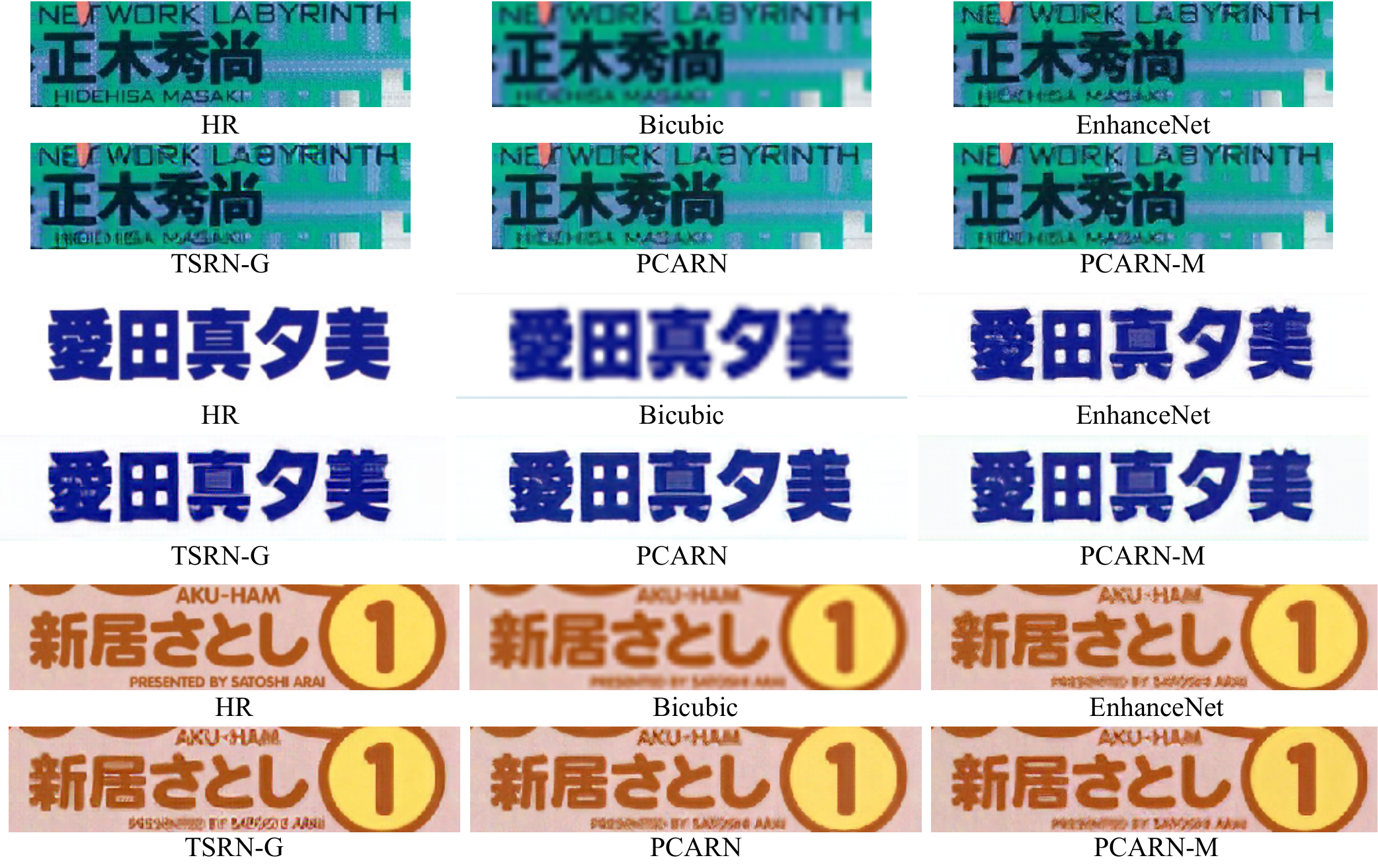}
\caption{\textbf{Visual comparison on text image.} We compare perception-based methods that have similar complexity to ours on comic book dataset, manga109~\cite{matsui2017sketch} ($\times$4 scale).}
\label{fig:comp_manga}
\end{figure*}

\section{Results}
We compare to following methods: SRCNN~\cite{srcnn2014}, MSLapSRN~\cite{MSLapSRN}, FEQE~\cite{vu2018fast}, SRResNet~\cite{srgan}, SRGAN~\cite{srgan}, EnhanceNet~\cite{enhancenet} (shortly ENet), TSRN\footnote{We choose TSRN-G model since it performs better.}~\cite{TSRN} and RankSRGAN~\cite{zhang2019ranksrgan}.
We categorize these as distortion- and perception-based by the use of perception loss (\textit{e.g.,} adversarial~\cite{goodfellow2014generative} or VGG~\cite{johnson2016perceptual}).
To measure the quality of the reconstructed images, we use both full-reference and non-reference quality assessments.
The former one calculates the score by comparing the SR and the reference ground-truth (HR) counterparts while the latter predicts the quality of the images as perceived by a human.
For full-reference image quality assessment, we utilize PSNR, SSIM~\cite{ssim}, LPIPS~\cite{zhang2018perceptual} and NIQE~\cite{mittal2012making}, while we employ  NIMA~\cite{talebi2018nima}, Ma~\cite{ma2017learning} and PI~\cite{blau2018perception} for the non-reference manner.

\subsection{Quantitative Comparison}

\Tref{table:benchmark_fr} depicts the quantitative comparisons for the $\times$4 scale dataset with full-reference quality measures including both pixel-based (PSNR and SSIM) and perception-based metrics (LPIPS and NIQE).
Except FEQE~\cite{vu2018fast}, PCARN and PCARN-M are the models with one of the least number of operations, \textit{i.e.,} MultAdds.
All the distortion-based methods show lower perceptual scores contrary to the superior PSNR and SSIM. This is because the pixel loss is not aligned with human perception.
For those which fall into the perception-based category, SRGAN, ENet, and TSRN-G have an identical number of parameters and MultAdds since they design the generator based on SRResNet~\cite{srgan}.
In contrast, our method shows the best performance on the LPIPS metric using a small number of MultAdds (120.G vs. 90.9G).
In detail, with the limited resources, PCARN surpasses all the competitors in LPIPS, and shows superior performance on most of the benchmark datasets in the NIQE metric.
Furthermore, the PCARN-M, which uses only one-fourth of both parameters and operations shows comparable scores to the state-of-the-art methods.
FEQE is the most lightweight network in terms of both parameter and operation, however, the SR quality of this model is far inferior to the other competitors.
We would also like to emphasize that among the perception-based models, the proposed PCARN family achieves the best performance for all benchmark datasets.
Such observation shows that our models achieve a good balance between perception and distortion, which is the ultimate goal of the SR algorithm~\cite{blau2018perception}.

\begin{figure}[t]
\centering
\begin{subfigure}[b]{0.99\linewidth}
    \centering
	\includegraphics[width=\linewidth]{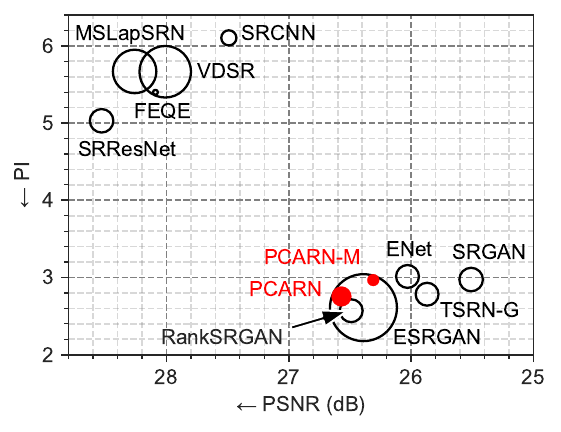}
\end{subfigure}
\\
\begin{subfigure}[b]{\linewidth}
    \centering
	\includegraphics[width=\linewidth]{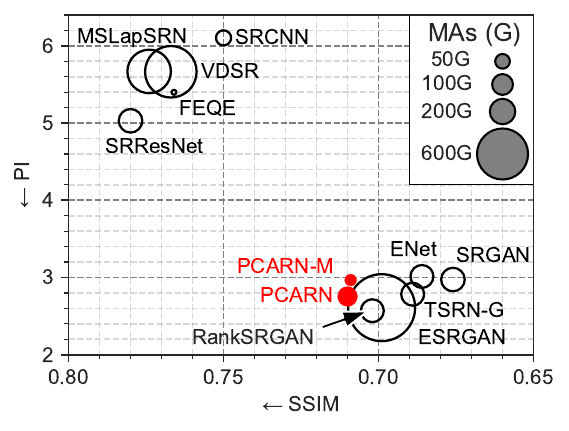}
\end{subfigure}
\caption{\textbf{Perception-distortion trade-off comparison.} We plot our proposed models with state-of-the-art methods on both perception (PI) and distortion (PSNR, SSIM) measures simultaneously in Set14 scale $\times$4 benchmark dataset. The size of the bubble indicates the number of operations,\textit{i.e.,} MultAdds.}
\label{fig:perception-distortion}
\end{figure}

In \Tref{table:benchmark_nr}, we compare the methods with non-reference quality assessments (Ma, PI, and NIMA).
As similar to the previous comparison (\Tref{table:benchmark_fr}), our methods show solid SR performance with fewer operations.
In particular, for all the benchmark metrics, PCARN achieves comparable to or better results than the perception-based counterparts, and PCARN-M shows reasonable performance as well.
We also visualize the perception-distortion trade-off comparison (\Fref{fig:perception-distortion}) as shown in Blau \textit{et al.}~\cite{blau2018perception}.
Although our proposed networks are not the most outstanding ones in terms of perceptual index (PI) and RankSRGAN~\cite{zhang2019ranksrgan} is the best, we achieve a good balance point between the distortion (PSNR and SSIM) and perception (PI).
We conjecture that it is because RankSRGAN relies on ranker loss which facilitates the perceptual metrics into their network training phase.
Overall, we can say that both our PCARN and RankSRGAN are admissible, as they are not dominated by any other algorithm compared in \Fref{fig:perception-distortion}.
That is, PCARN is superior to RankSRGAN in terms of distortion while RankSRGAN achieves a better score in the perceptual index.
Similarly, our PCARN and ESRGAN~\cite{wang2018esrgan} are also comparable, however, our method only uses ten times smaller MultAdds than ESRGAN.
Also note that, PCARN and PCARN-M are the most lightweight networks compared to the others that belong to the perception-driven group (\textit{i.e.,} lower PSNR, SSIM, and PI).

We also report the perception scores on various scale factors (\Tref{table:ours_perception}).
Our models are capable of processing multiple scales with a single network.
Unlike ours, a network without multi-scale training can only restore a specific scale, which limits the generalization ability to unseen degradation.
However, our methods enjoy such generalization by using the multi-scale training strategy.

\subsection{Visual Comparison}
\begin{figure}[t]
\centering
\includegraphics[width=\linewidth]{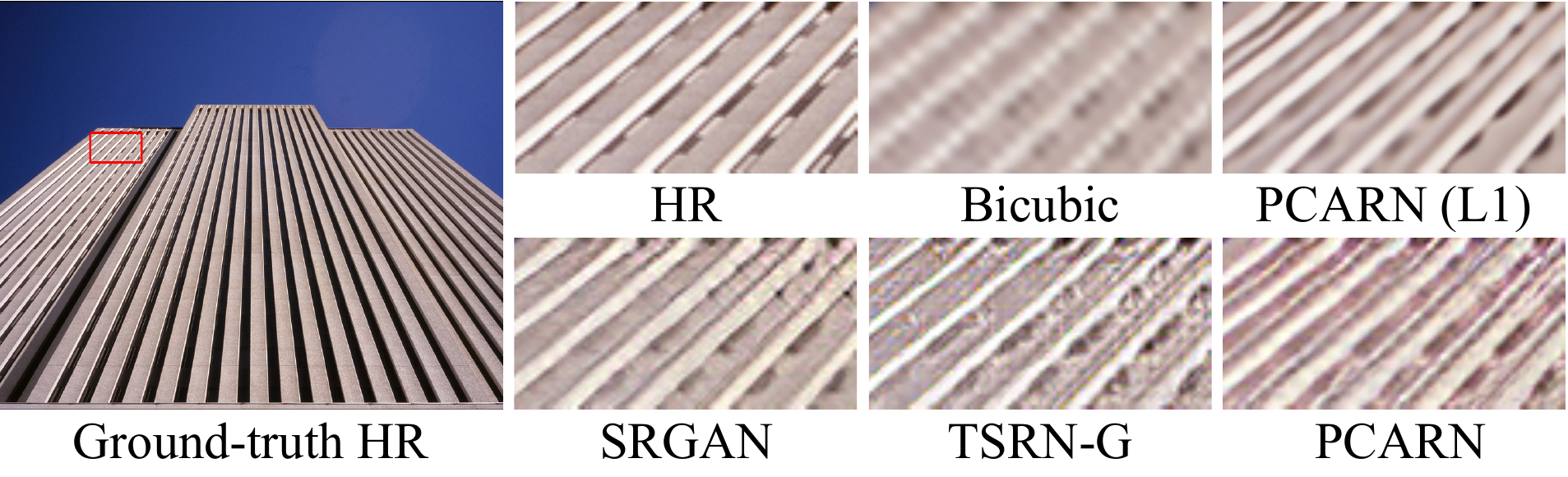}
\caption{\textbf{A failure case.} Our perception-based model is not able to reconstruct the details without sufficient information when constructing a dense structure.}
\label{fig:failure}
\end{figure}

In \Fref{fig:comp_perception}, we illustrate the qualitative comparisons on the various $\times$4 scale datasets.
It can be seen that our models work better than others and accurately reconstruct not only the linear patterns but produce more photo-realistic textures, such as the mane of the lion and pebbles on the ground.
Moreover, the proposed networks also generate cleaner outputs while other perception-based methods suffer visual artifacts.
To investigate how our models can generate SR image from different domains, we examine the visual comparison on text images using manga109~\cite{matsui2017sketch} (\Fref{fig:comp_manga}).
Since humans can easily distinguish high-frequency details, it is important to adequately recover the edge region in this task. 
Our method effectively restores various texts, even those with very small fonts that are barely recognizable on the bicubic results.
GAN-based PCARN can produce sharp and realistic images. However, for some dense structures as in \Fref{fig:failure}, it generates undesirable artifacts, unlike the L1 loss-based PCARN.
We suspect that this is a common limitation shared by most, if not all, GAN-based algorithms.

\subsection{Execution time}
\label{sec:time}

\begin{figure*}[t]
	\centering
	\begin{subfigure}[b]{0.49\linewidth}
		\includegraphics[width=\linewidth]{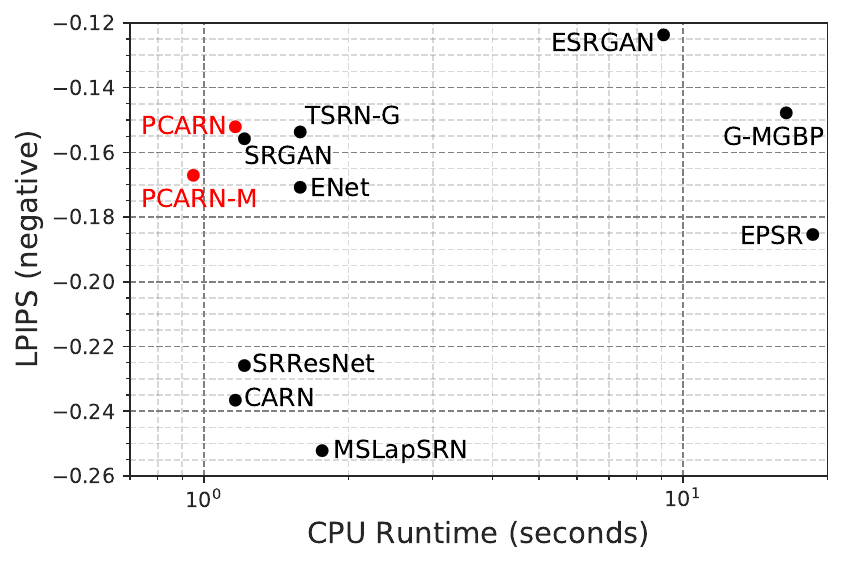}
	\end{subfigure}
	\begin{subfigure}[b]{0.49\linewidth}
		\includegraphics[width=\linewidth]{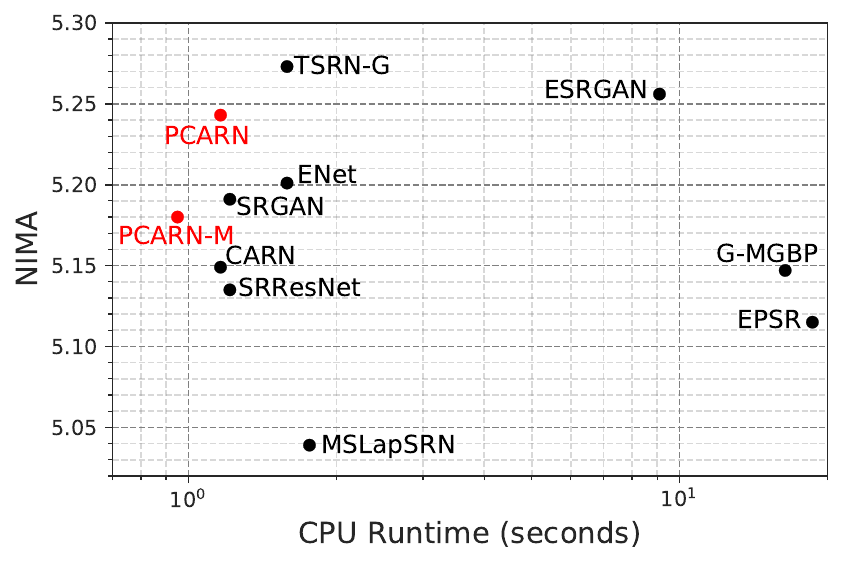}
	\end{subfigure}

	\begin{subfigure}[b]{0.49\linewidth}
		\includegraphics[width=\linewidth]{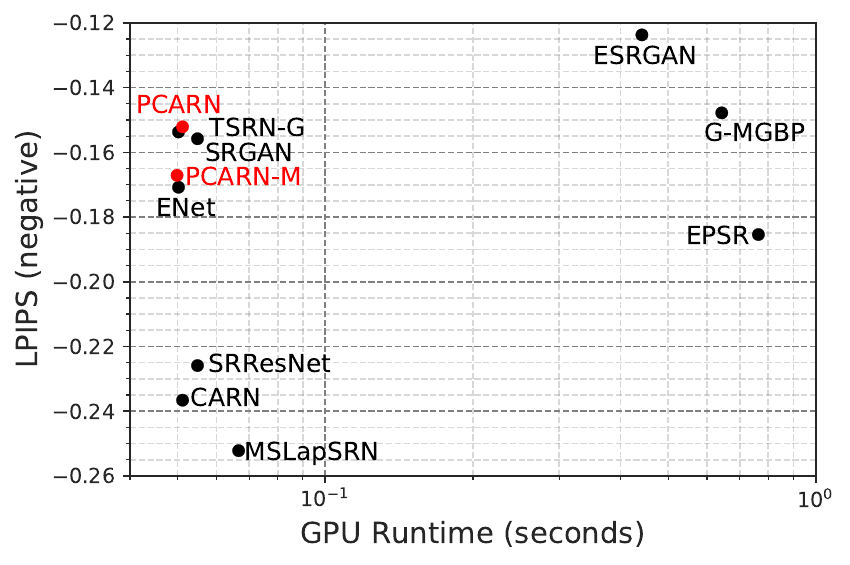}
	\end{subfigure}
	\begin{subfigure}[b]{0.49\linewidth}
		\includegraphics[width=\linewidth]{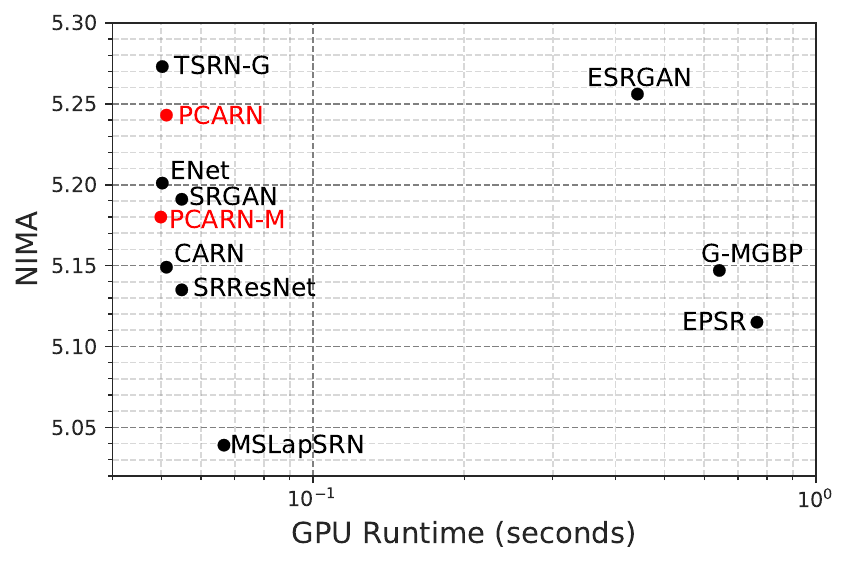}
	\end{subfigure}
	\caption{\textbf{Execution time on the CPU/GPU environments.} We measure the the runtime on CPU and GPU settings with LPIPS and NIMA score. We use negative LPIPS to match the direction of y-axis to the NIMA.}
\label{fig:runtime}
\end{figure*}

While MultAdds can reflect the heaviness of the model well, there still exists a misalignment between the true execution times, especially when the models are run on GPU~\cite{shufflenetv2}.
In order to investigate the efficiency in the real devices, we evaluate the inference runtime with other state-of-the-art networks (\Fref{fig:runtime}).
In this benchmark, we compare computationally-heavy models as well (ESRGAN~\cite{wang2018esrgan}, G-MGBP~\cite{G-MGBP} and EPSR~\cite{vasu2018analyzing}).
For a fair comparison, we perform inference on the same machine (NVIDIA TITAN X GPU).
To calculate inference time, we use a resolution of 320$\times$180 for the LR input so that the network generates a 720p (1280$\times$720) SR image.

For CPU execution (top row in \Fref{fig:runtime}), the speed of PCARN is faster than the other methods such as SRGAN and SRResNet, while it produces a better result and is comparable with the EPSR and G-MGBP.
Our PCARN-M is the fastest, while on a par with the heavy models.
Such illustration is also reflected by the NIMA metric.
The PCARN and PCARN-M can obtain good results at a relatively low computational cost.
However, unlike the results on CPU, our methods do not show such improvement on the GPU (bottom row in \Fref{fig:runtime}). In fact, our models show slightly worse execution times than the ENet and TSRN-G.
The reason is mainly due to the distinct characteristic of CPU and GPU.
For example, as empirically proved in \cite{shufflenetv2}, memory fragmentation reduces the parallelism which worsens the GPU speed a lot.
In our case, the cascading mechanism hinders GPU parallelism so that both PCARN and PCARN-M have fewer advantages on GPU.
Furthermore, group convolution is not implemented in a GPU-friendly manner, diminishing the speed gap between PCARN and PCARN-M.

\section{Conclusion}
In this work, we proposed the model with a cascading scheme for fast and accurate SR.
The core idea is adding multiple cascading connections starting from each intermediary layer to the others on local and global levels.
In addition, we enhance our model by using a multi-scale discriminator and achieved improved SR quality over the recent models that have complexity on par with ours.
All the experiments have been conducted on the SR task, but we expect that our work can potentially be applied to generic image-to-image translation or image enhancement as well.

While our methods achieve efficiency, there are remaining issues such as improving the usability and robustness of our models.
First of all, the GPU execution time is different from the time taken under the CPU setting (\Sref{sec:time}), despite the decreased number of MultAdds.
This phenomenon comes up because of the discrepancy between the MultAdds and the actually-measured benchmark time.
While MultAdds can reflect the inference speed on the CPU, but for the GPU, there are many uncounted operations that MultAdds does not account for.
Therefore, our future goal is to improve our framework and build a GPU-friendly network by carefully modifying our modules and convolution.

Another issue is related to the limitation of the network itself, which manifests in failures such as the bad reconstruction of dense and small textures (\Fref{fig:failure}).
For the future direction, we hope to use ideas from example-based SR~\cite{yang2019example} or non-local neural networks~\cite{liu2018non} so that it can effectively enhance severely distorted regions.

\smallskip
\noindent\textbf{Acknowledgement.}
This research was supported by the MSIT(Ministry of Science and ICT), Korea, under the ITRC(Information Technology Research Center) support program(IITP-2021-2018-0-01431), and under Grant 2021-0-02068 (Artificial Intelligence Innovation Hub), supervised by the IITP(Institute for Information \& Communications Technology Planning \& Evaluation). N.A. was also supported by the BK21 FOUR program of the NRF of Korea funded by the Ministry of Education (NRF5199991014091).

{
\small
\bibliographystyle{./cvpr/ieee_fullname}
\bibliography{}
}

\end{document}